\documentclass[11pt,a4paper]{article}

\usepackage[utf8]{inputenc}   
\usepackage[T1]{fontenc}          
\usepackage{graphicx}                   
\usepackage{float}                      
\usepackage{geometry}                   
\usepackage{hyperref}                   
\usepackage{setspace}                   
\usepackage{color}                      
\usepackage[table]{xcolor}              
\usepackage{fancyhdr}                   
\usepackage{fourier-orns}               
\usepackage[usenames,dvipsnames]{pstricks}

\hypersetup{                            
        colorlinks=true,                
        breaklinks=true,                
        urlcolor=black,                 
        linkcolor=black,                
        pdftitle={},                    
        pdfauthor={O. Bailleux},        
        pdfproducer={O. Bailleux},      
        pdfsubject={PPC},               
        pdftoolbar=true,                
        pdfmenubar=true,                
        pdfstartview={FitH},            
}

\sffamily                               
\title{\vspace{-1cm}On the \textsc{cnf} encoding of cardinality constraints and beyond}

\normalsize\author{Olivier Bailleux, Université de Bourgogne}

\usepackage{amssymb}

\begin{document}

\sloppy 

\rfoot{\small page \thepage / \pageref{pageend}}

\maketitle 



\begin{spacing}{1.05}

\begin{abstract}
In this report, we propose a quick survey of the currently known techniques for encoding a Boolean cardinality constraint into a \textsc{cnf} formula, and we discuss about the relevance of these encodings.
We also propose models to facilitate analysis and design of \textsc{cnf} encodings for Boolean constraints.
\end{abstract}

\section{Introduction}

In this report, we propose a quick survey of the currently known techniques for encoding a Boolean cardinality constraint into a \textsc{cnf} formula, and we discuss about the relevance of these encodings.

A Boolean cardinality constraint can be denoted $\leq k(x_1,\ldots,x_n)$, meaning $\sum_{i=1}^n{x_i} \leq k$. A \textsc{cnf} formula is a disjunction of clauses, where each clause is a conjunction of literals, where each literal is either a propositional variable or a negated propositional variable. For convenience, such a formula can be represented as a set of clauses, where each clause is a set of literals.

Given a set $V=\{v_1,\ldots,v_n\}$ of propositional variables, a partial truth assignment on $V$ is a set $I$ of literals such that for any $w\in I$, either $w$ or $\overline w$ is in $V$, and $\overline w \notin I$. A complete truth assignment on $V$ is a partial assignment on $V$ such that for any $v \in V$, either $v$ or $\overline v$ is in $I$.
Given a truth assignment $I$ and a formula $\sigma$, $\sigma|_I$ denotes $\sigma \wedge_{w\in I} (w)$.

A formula $\sigma$ is said to \emph{encode} of a given constraint $q(x_1,\ldots,x_n)$ if and only if, for any complete truth assignment $I$ on $V = \{x_1,\ldots,x_n\}$, $\sigma|_I$ is satisfiable if and only if $I$ satisfies $q$. 
It is said to be a \texttt{pac} (like \emph{propagating arc consistency}) encoding if and only if, given any partial truth assignment $I$, applying unit propagation on $\sigma|_I$ fixes the same variables of $V$ as restoring arc consistency on $q$.
It is said to be a \texttt{pic} (like \emph{propagating inconsistency}) encoding if and only if, given any partial truth assignment $I$, applying unit propagation on $\sigma|_I$ produces the empty clause if and only if $I$ falsifies $q$.

\section{Existing encodings\label{surway}} 

Existing encodings can be roughly classified into two (overlapping) categories : the ones which are based on a bit counter coupled with a comparator, and the encodings dedicated to the pseudo-Boolean constraints, i.e., constraints of the form $a_1 x_1+\cdots+a_n x_n \leq b$, where $a_1,\ldots,a_n,b$ are positive integers and $x_1,\ldots,x_n$ are propositional literals.

\subsection{Encodings based on bit counters}

These encodings are based on a Tseitin transformation \cite{tseitin68} of a circuit including a bit counter cascaded with a comparator. Two approaches has been proposed, respectively based on a unary and a binary representation of the counter output.

\subsubsection{Binary representation}

These encodings use binary adders and comparators. Warner introduced such an approach in \cite{warner68} for translating a pseudo-Boolean constraints into a \textsc{cnf} formula. The proposed solution can be simplified in the particular case of cardinality constraints. The size of the obtained formula is linearly related to the number of variables in the input constraint, as well as to  the number of auxiliary variables. 
Another architecture is proposed in \cite{Sinz05towardsan}, where the bit counter is organized as a tree of binary adders. The size of the resulting formula and the number of required auxiliary variables are $\Theta(n)$\footnote{ When applicable, we prefer we prefer to use the $\Theta$ notation rather than the $O$ notation, because the latter is only an upper bound. For example, $n \log n = O(2^n)$.}.
These encodings are known to be neither \texttt{pac}, nor \texttt{pic}.

\subsubsection{Unary representation}

By adopting a unary representation of the output of the bit counter, which incidentally makes obvious the comparison stage, we obtain encodings techniques which produce larger formulae, but allow unit propagation to perform more deductions.

This was shown for the first time in \cite{DBLP:conf/cp/BailleuxB03}, where a \texttt{pac} encoding requiring $\Theta(n^2)$ clauses and $\Theta(n \log n)$ auxiliary variables is presented. The unary bit counter of $n$ inputs is designed as an association of two bit counters of $n/2$ inputs coupled with a unary adder.

Another architecture is proposed in \cite{Sinz05towardsan}, where the bit counter is shaped as a sequential association of unary adders. The resulting \texttt{pac} encoding requires $\Theta(nk)$ clauses and auxiliary variables.

As shown in \cite{DBLP:journals/jsat/EenS06}, the bit counter can also be done thanks to a sorting network. This approche allows to produce a \textsc{cnf} formula with $\Theta(n \log^2 n)$ clauses and auxiliary variables.

All these encodings are \texttt{pac}, then \texttt{pic}. As far as we know, no criteria have been proposed to decide between them.

\subsection{pseudo-Boolean encodings}

Because a Boolean cardinality constraint is a special case of pseudo-Boolean constraint, any encoding dedicated to pseudo-Boolean constraints can be used with pseudo-Boolean ones. Excluding approaches producing a formula of exponential size and the ones that have been mentioned above, this covers three techniques, namely the \texttt{bdd} encoding presented in \cite{DBLP:journals/jsat/BailleuxBR06}, and the two encodings presented in \cite{DBLP:conf/sat/BailleuxBR09}, namely \texttt{lpw} and \texttt{gpw}.

The \texttt{bdd} based encoding presented in \cite{DBLP:journals/jsat/BailleuxBR06} is a \texttt{pac} \textsc{cnf} encoding of pseudo-Boolean contraints that can produce an exponential size formula in the worst case. But with a cardinality constraint as input, the size of the resulting formula is $O(nk)$, which is competitive with other encodings presented above.

In contrast, the \texttt{gpw} encoding introduced in \cite{DBLP:conf/sat/BailleuxBR09} presents no interest because with a cardinality constraint as input, it falls down to the encoding presented in \cite{DBLP:conf/cp/BailleuxB03}. Finally, the \texttt{lpw} encoding requires a unary bit counter for each
variable of the input constraint. It produces a formula of size $\Theta(kn^2)$, which is  somewhat prohibitive.

\section{Discussion}

\subsection{Binary versus unary representation}

In all the encodings that we have reviewed, there is an implicit or explicit calculation of the number of variables that are fixed to 1 among the input variables. With the encodings based on binary arithmetic, this calculation requires that all the input variables are assigned because the Boolean functions related to each bit of the binary representation are not monotonic with respect to the number of input variables that are fixed to one. For example, the lower bit of this representation depends only on the parity of the input cardinality, then alternatively changes each time this input cardinality increases. This structurally prevents some propagations when some input variables are not assigned, even if there are enough input variables fixed to 1 for falsifying the constraint. As a consequence, some inconsistencies cannot be detected by unit propagation alone.

On the other hand, the encodings based on unary arithmetic allow unit propagation to calculate the input cardinality even when some input variables are not assigned. This is made possible by the monotonicity of the functions related to each bit of the unary representation of the input cardinality. This deserves an explanation, given Sections \ref{monotonic} and \ref{complexity}.

\subsection{Filtering functions \label{monotonic}} 

Restoring arc consistency of a Boolean cardinality constraint -- but also other Boolean constraints, like the pseudo-Boolean ones -- reduces to compute functions which map $\{0,1,*\}^n$ to $\{0,1,*\}$, where the symbol $*$ means that a variable is not assigned.

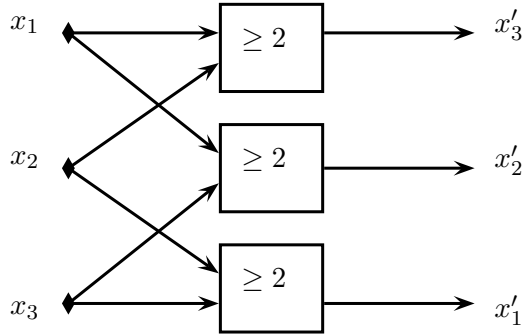
\begin{figure}[h]
\begin{center}
\scalebox{1} 
{
\begin{pspicture}(0,-2.2)(7.3628125,2.2)
\psdots[dotsize=0.2,dotstyle=diamond*](0.9809375,1.8)
\psdots[dotsize=0.2,dotstyle=diamond*](0.9809375,0.0)
\psdots[dotsize=0.2,dotstyle=diamond*](0.9809375,-1.8)
\psline[linewidth=0.04cm,arrowsize=0.12cm 2.0,arrowlength=1.4,arrowinset=0.4]{->}(0.9809375,1.8)(2.9809375,1.8)
\psframe[linewidth=0.04,dimen=outer](4.3809376,0.6)(2.9809375,-0.6)
\psframe[linewidth=0.04,dimen=outer](4.3809376,2.2)(2.9809375,1.0)
\psframe[linewidth=0.04,dimen=outer](4.3809376,-1.0)(2.9809375,-2.2)
\psline[linewidth=0.04cm,arrowsize=0.12cm 2.0,arrowlength=1.4,arrowinset=0.4]{->}(0.9809375,0.0)(2.9809375,1.4)
\psline[linewidth=0.04cm,arrowsize=0.12cm 2.0,arrowlength=1.4,arrowinset=0.4]{->}(0.9809375,0.0)(2.9809375,-1.4)
\psline[linewidth=0.04cm,arrowsize=0.12cm 2.0,arrowlength=1.4,arrowinset=0.4]{->}(0.9809375,1.8)(2.9809375,0.2)
\psline[linewidth=0.04cm,arrowsize=0.12cm 2.0,arrowlength=1.4,arrowinset=0.4]{->}(0.9809375,-1.8)(2.9809375,-1.8)
\psline[linewidth=0.04cm,arrowsize=0.12cm 2.0,arrowlength=1.4,arrowinset=0.4]{->}(0.9809375,-1.8)(2.9809375,-0.2)
\psline[linewidth=0.04cm,arrowsize=0.12cm 2.0,arrowlength=1.4,arrowinset=0.4]{->}(4.3809376,1.8)(6.3809376,1.8)
\psline[linewidth=0.04cm,arrowsize=0.12cm 2.0,arrowlength=1.4,arrowinset=0.4]{->}(4.3809376,-1.8)(6.3809376,-1.8)
\psline[linewidth=0.04cm,arrowsize=0.12cm 2.0,arrowlength=1.4,arrowinset=0.4]{->}(4.3809376,0.0)(6.3809376,0.0)
\usefont{T1}{ptm}{m}{n}
\rput(0.40234375,1.91){$x_1$}
\usefont{T1}{ptm}{m}{n}
\rput(0.40234375,0.11){$x_2$}
\usefont{T1}{ptm}{m}{n}
\rput(0.40234375,-1.89){$x_3$}
\usefont{T1}{ptm}{m}{n}
\rput(6.842344,-1.89){$x_1'$}
\usefont{T1}{ptm}{m}{n}
\rput(6.842344,0.11){$x_2'$}
\usefont{T1}{ptm}{m}{n}
\rput(6.842344,1.91){$x_3'$}
\usefont{T1}{ptm}{m}{n}
\rput(3.5823438,1.71){$\geq 2$}
\usefont{T1}{ptm}{m}{n}
\rput(3.5823438,0.11){$\geq 2$}
\usefont{T1}{ptm}{m}{n}
\rput(3.5823438,-1.49){$\geq 2$}
\end{pspicture} 
}
\caption{Filtering network for $\leq 2(x_1,x_2,x_3)$\label{fig1}}
\end{center}
\end{figure}

Regarding the constraint $\leq k(x_1,\ldots,x_n)$, these \emph{filtering functions} are of the form  $f_{q,m}$ such that if the number of input values set to $1$ among $y_1,\ldots,y_m$ is at least $q$ then $f_{q,m}(y_1,\ldots,y_m)=0$, else $f_{ q,m}(y_1,\ldots,y_m)=*$. For example, the filtering function related to the input variable $x_1$ is $f_{k,n-1}(x_2,\ldots,x_n)$ because, accordingly to the principle of arc consistency, $x_1$ must be fixed to 0 if there are $k$ other input variables assigned to 1. If there are \emph{more} than $k$ input variables assigned to 1, and if the value of each $x_i$ is determined by the related filtering function, then a contradiction occurs, i.e. at least one of the input variable is fixed both to 1 (by hypothesis) and to 0 (by the related filtering function). Therefore, any \texttt{pac} encoding of a Boolean cardinality constraint $\leq k(x_1,\ldots,x_n)$ explicitly or implicitly allows unit propagation to compute the filtering functions related to each input variable $x_i$, and any encoding allowing unit propagation to  compute
these functions is \texttt{pac}.

As a example, Figure \ref{fig1} shows the implicit filtering functions for the constraint $\leq 2(x_1,x_2,x_3)$. Each of these functions maps $\{0,1,*\}^2$ to $\{0,*\}$, with output value 0 if and only if its two inputs are set to 1. Each output $x_i'$ represents the value of the variable $x_i$ after the propagation is done.
The clause $(\overline x_1 \vee \overline x_2 \vee \overline x_3)$ allows unit propagation to achieve all these computations.

Without lost of generality, we can consider that the filtering functions have codomain $\{1,*\}$, because
\begin{enumerate}
\item 
any filtering function $f$ with codomain $\{0,1,*\}$ can be decomposed into two simplified filtering functions $f_0, f_1$ with domains $\{0,*\}$ and $\{1,*\}$, respectively;

\item
to any filtering function $f$  can be associated a filtering function $\overline f$  such that for any suitable input assignment $I$, if $f(I)=*$ then $\overline f(I)=*$, else $\overline f(I) = 1-f(I)$;

\item
any formula $\phi$ computing a  filtering function $f$ with output variable $s$ can compute $\overline f$ with output variable $t$ by adding the clauses $(s \vee \overline t) \wedge (\overline s \vee t)$;

\item
for any filtering function $f$, if the formulae $\phi_0, \phi_1$ compute $f_0, f_1$ with output variables $s_0, s_1$ (assuming without lost of generality that $\phi_0, \phi_1$ share no variable except the input ones) then the formula $\phi_0 \wedge \phi_1 \wedge (s_0 \vee \overline s) \wedge (\overline s_1 \vee s)$ computes $f$ with output variable $s$. 
\end{enumerate}

Why these filtering functions cannot be propagated through a binary representation ? As an example, let us consider the constraint $\leq 2(x_1,x_2,x_3,x_4)$. In any encoding based on a binary representation, by definition, there are three auxiliary variables $u_2,u_1,u_0$ representing the binary value of the number of input variables fixed to 1. These variables link the output of the bit counter (whatever its architecture) with the comparator. Now, suppose the two input variables $x_1,x_2$ are fixed to 1, and the two other ones, i.e., $x_3,x_4$, are not fixed.
This means that the input cardinality could be 2,3, or 4, hence, in binary, 010, 011, or 100. Each of the variables $s_2,s_1,s_0$ could take either the value 0 or 1, depending on the further values of $x_3,x_4$. While these variables are not fixed, nothing can be inferred regarding the values of $s_2,s_1,s_0$. Then the comparator cannot "know" that $x_3,x_4$ must be set to 0. Worse, if three input variables are fixed to 1 and the other one is not fixed, each of the variable $s_2,s_1,s_0$ can be potentially fixed to 0 or 1, then the inconsistency cannot be detected.

Finally, remark that the \texttt{bdd} encoding can be considered as based on a unary encoding, because each of the nodes of the underlying decision diagram is related to a  filtering function in the sense described above.

\section{Complexity issues \label{complexity}}

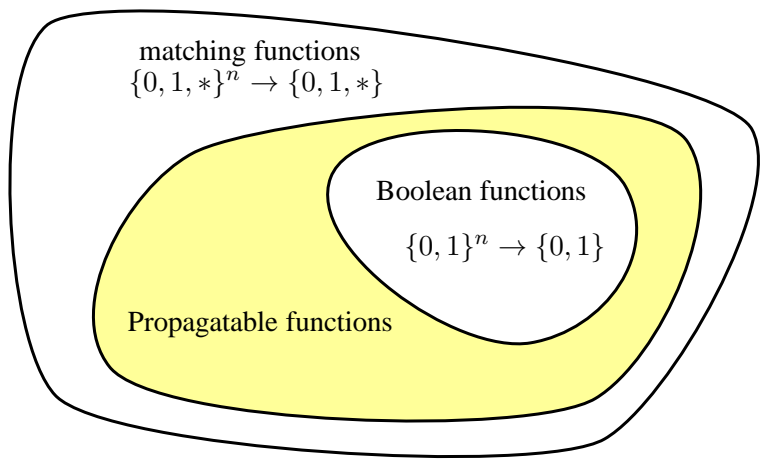
\begin{figure}[h]
\begin{center}
\scalebox{1} 
{
\begin{pspicture}(0,-3.3514593)(10.619721,3.3514595)
\definecolor{color72b}{rgb}{1.0,1.0,0.6}
\psbezier[linewidth=0.04,fillstyle=solid,fillcolor=color72b](2.688795,0.96865785)(3.568795,1.4486579)(8.688795,1.9886578)(9.268795,1.1286579)(9.848795,0.2686579)(8.905859,-1.7761322)(8.048795,-2.291342)(7.191731,-2.806552)(2.2104852,-2.6546054)(1.5887951,-1.8713421)(0.9671048,-1.088079)(1.8087951,0.4886579)(2.688795,0.96865785)
\psbezier[linewidth=0.04,fillstyle=solid](4.5287952,0.64865786)(4.8818326,1.584267)(7.9713264,1.4073067)(8.448795,0.52865785)(8.926264,-0.34999093)(8.17479,-1.3845669)(7.188795,-1.5513421)(6.2028003,-1.7181174)(4.1757574,-0.2869512)(4.5287952,0.64865786)
\usefont{T1}{ptm}{m}{n}
\rput(6.5395765,0.47865787){Boolean functions}
\usefont{T1}{ptm}{m}{n}
\rput(6.850201,-0.28134212){$\{0,1\}^n \rightarrow \{0,1\}$}
\usefont{T1}{ptm}{m}{n}
\rput(3.5995762,-1.3013421){Propagatable functions}
\usefont{T1}{ptm}{m}{n}
\rput(3.4609826,2.298658){matching functions}
\usefont{T1}{ptm}{m}{n}
\rput(3.5502014,1.8986579){$\{0,1,*\}^n \rightarrow \{0,1,*\}$}
\psbezier[linewidth=0.04](0.7487951,2.6686578)(1.4975902,3.3314595)(9.657869,2.1708305)(10.128795,1.2886579)(10.599721,0.40648517)(9.034753,-2.3312247)(8.168795,-2.8313422)(7.3028374,-3.3314595)(1.5562278,-2.97759)(0.8687951,-2.251342)(0.18136236,-1.525094)(0.0,2.0058563)(0.7487951,2.6686578)
\end{pspicture} 
}
\caption{Boolean, matching and propagatable functions\label{fig2}}
\end{center}
\end{figure}

In this section, we ask different questions about the complexity of \texttt{pac} and \texttt{pic} \textsc{cnf} encodings for Boolean cardinality constraints as well as for other kind of constraints. First, let us recall that any Boolean function can be computed using unit propagation under the assumption that the input value is represented as a complete truth assignment of the input variables. This is due to the fact that any Boolean function can be computed thanks to a Boolean circuit, and that the behavior of any Boolean circuit with $n$ nodes can be simulated by applying unit propagation on a formula of $O(n)$ clauses.
But not all functions mapping $\{0,1,*\}^n$ to $\{0,1,*\}$ -- that we propose to call \emph{matching functions} -- can be computed in this way. For example, the function $h$ that maps $\{0,1,*\}$ to $\{0,1,*\}$ such that $h(0)=0, h(1)=1, h(*)=0$ cannot. Informally speaking, unit propagation cannot test whether a variable is assigned or not. We propose to call \emph{propagatable functions} the matching functions that can be computed thanks to unit propagation (see Figure \ref{fig2}). 

Clearly, the propagatable functions are the matching functions that are monotonic with respect to the following order: $* \preccurlyeq *$, $* \preccurlyeq 1$, $* \preccurlyeq 0$, $0 \preccurlyeq 0$, $1 \preccurlyeq 1$, $(x_1, \ldots,x_n) \preccurlyeq (y_1, \ldots,y_n)$ if and only if $x_i \preccurlyeq y_i, 1 \leq i \leq n$.
It follows that for any Boolean constraint $q(x_1,\ldots,x_n)$ and any $ 1\leq i \leq n$, the filtering function related to $x_i$ is a propagatable function, because if the value of $x_i$ can be inferred whereas some other input variables are not fixed, then this value of $x_i$ holds whatever the values of these input variables. Thus, the complexity of computing propagatable functions with unit propagation is a key concept in studying the \textsc{cnf} encoding of Boolean constraints.

Now, let us present the critical issues regarding the search for efficient encodings. The following questions are relevant for Boolean cardinality constraints, but can be generalized to other constraints on Boolean variables, such as pseudo-Boolean constraints.

\begin{enumerate}
\item 
The smallest known \texttt{pac} encoding for Boolean cardinality constraint is presented in \cite{DBLP:journals/jsat/EenS06}. This is actually a pseudo-Boolean to \textsc{cnf} encoding which is not \texttt{pac} for any pseudo-Boolean input constraint, but which is \texttt{pac} in the particular case of cardinality constraints. The size of the output formula is $\Theta(n \log^2 n)$, which is better than $\Theta(kn)$ when $k=\Theta(n)$. 

Is there a smaller \texttt{pac} encoding? Is there a \texttt{pac} encoding which produce a formula of size $O(n)$? Is there a gap between the smaller \texttt{pac} encoding and the smaller encoding with binary representation?

\item
The preceding questions are about encodings of a whole cardinality constraints, which implicitly include the filtering functions related to each input variables. But what about the size complexity of computing (thanks to unit propagation) each filtering function ? Clearly, if each filtering function requires $\Theta(f(n))$ clauses, then size of the smallest \texttt{pac} encoding is  $O(n f(n))$, but not necessarily $\Omega(n f(n))$, because some parts of the output formula could be shared to compute several filtering functions.

The smallest known encodings for Boolean cardinality constraints allow to compute the underlying filtering functions with a formula of size $\Theta(n \log^2 n)$ (assuming $k=\Theta(n)$), i.e., the same size as for restoring arc consistency on the whole constraint. Is it possible to do better?
\end{enumerate}

\section{Concluding remarks and perspectives}

There are at least three research ways in the field of \textsc{cnf} encoding of Boolean (including cardinality) constraints. The first direction is the research for theoretical models that would facilitate the design and the analysis of encodings. The second one concerns the research for inference rules and filtering techniques that would allow \textsc{sat} solvers to achieve the same deductions with binary encodings as the current solvers do with unary encodings. And the last research area is a fine study of the respective inference powers and efficiencies of \textsc{sat} and pseudo-Boolean solvers regarding the problems which can be represented with the two formalisms.

\subsection{Designing and analysing \textsc{cnf} encodings \label{monotoneversusup}}

A way to design propositional encodings is to start from a Boolean circuit, then use a Tseitin transformation to produce the corresponding formula. Indeed, all the encodings presented in section \ref{surway} can be represented as Boolean circuits. This representation is suitable for designing correct encodings and for proving the correctness of encodings. But it does not model the behavior of unit propagation alone, especially when some variables are not assigned.

The way unit propagation computes a filtering function can  be simulated with a monotone Boolean circuit by representing each of the three possible values of any variable $u$ by two binary values $u^+, u^-$ such that $u=*$ is represented by $u^+=0,u^-=0$, $u=0$ is represented by $u^+=0,u^-=1$, and $u=1$ is represented by $u^+=1,u^-=0$. 
It is easy to see that any filtering function which can be computed with  a monotone circuit can also be computed by unit propagation with a  satisfiable \textsc{cnf} formula of the same size.
Conversely, under the assumption that the size of the clauses is bounded, any filtering function which can be computed with unit propagation on a satisfiable\footnote{ In fact, it suffices that the unit propagation can not produce a contradiction, so that the filtering function is fully defined on its domain. The filtering process will detect a local inconsistency when the result of the computation of the filtering function is in conflict with the initial value of some input variables.} \textsc{cnf} formula $\phi$ can also be computed with a monotone circuit of size linearly related to the size of $\phi$. A sketch of the proof is given in Annex \ref{cnftomonotone}.

Then there is a tight relation between the size of the smallest \textsc{cnf} formula computing a filtering function and monotone circuit complexity. Namely, this formula reduces to the smallest monotone circuit computing the related Boolean function.

\subsection{Improving filtering in \textsc{sat} solvers}

As mentioned before, the unassigned variables impact the expressive power of unit propagation. A possible way to overcome this problem is to "inform" unit propagation thanks to preassigned variables. For example, let us consider the constraint $\leq k(x_1,\ldots,x_n)$ where some variables are fixed to 1, some are fixed to 0, and the other are not assigned. We propose to achieve unit propagation under the assumption that unassigned variables are fixed to 0. Namely, these variables are considered as unassigned \emph{except for unit propagation}. If unit propagation fixes such a variable to 1, this is not considered as a conflict and the new value replaces the initial default one.

With this simple modification, which supposes to inform the \textsc{sat} solver of the default values of the involved variables, the binary based encodings for cardinality constraints become \texttt{pic}, allowing to increase the amount of deductions performed by the solver.
Thanks to such an informed unit propagation rule, we can expect more compact \texttt{pic} and \texttt{pac} encodings.

\subsection{\textsc{sat} versus pseudo-Boolean solvers}

Given what we said in this report, translating a Boolean cardinality constraint -- and more generally a pseudo-Boolean constraints -- in propositional formula is not obvious. There are many ways to proceed, with their advantages and disadvantages. On the other hand, Translating a clause into a pseudo-Boolean or cardinality constraint is immediate.

Therefore, it is questionable whether it is appropriate to use a SAT solver to deal with problems specified using both clauses and pseudo-Boolean constraints. Is it not possible to achieve, if indeed it does not already exist, a pseudo-Boolean solver that would be as efficient as a \textsc{sat} solver when dealing with clauses, and at least as efficient as a \textsc{sat} solver associated with a \textsc{cnf} encoding when dealing with other pseudo-Boolean constraints?

The question deserves a comparative study of existing \textsc{sat} and pseudo-Boolean solvers, with the same problem instances and all the known encodings of pseudo-Boolean and cardinality constraints. If it turns out that in some cases \textsc{sat} solvers are better, it will be relevant to investigate the reason for such a difference: learning strategy, branching heuristic, filtering efficiency... in order to be able to design a pseudo-Boolean solver which covers efficiently the scope of the \textsc{sat} solvers.

\appendix

\section{Reducing a \textsc{cnf} formula to a monotone circuit\label{cnftomonotone}}
To each filtering function $f(x_1,\ldots,x_n)$ with domain $\{1,*\}$ can be associated a Boolean function $f_\mathbb{B}(x_1^+, x_1^-,\ldots,x_n^+, x_n^-)$ with the convention introduced Section \ref{monotoneversusup}. Our aim is to prove that if $f$ can be computed using unit propagation on a satisfiable formula $\phi$ of size $m$, then $f_\mathbb{B}$ can be computed by a monotone Boolean circuit of size $O(m)$.

For any variable $x$, let us define $\delta(x) = x^+$ and $\delta(\overline x) = x^-$.

Without lost of generality, we want to prove that for any function $f$ mapping $\{0,1,*\}^n$ to $\{1, *\}$, if there is a \textsc{cnf} formula $\phi$ allowing unit propagation to compute $f$, then there is a monotone circuit which computes  $f_\mathbb{B}$.

Such a circuit can be built based on the following principle: all the deductions unit propagations can do regarding a given literal $w$, and then the corresponding Boolean variable $\delta(w)$, involve only the clauses containing $w$. Let $Q$ be the set of these clauses.

Clearly, the following part of circuit $C_w$ computes the value of $\delta(w)$ with respect to the variables which it depends.

\[
\delta(w) = \mathtt{or}(\{ \mathtt{input}(\delta(w)),\mathtt{and}(\{\delta(\overline l), l\in c, l \neq w\}),c\in Q\})
\]

\begin{figure}[h]
\begin{center}
\scalebox{1} 
{
\begin{pspicture}(0,-1.1178125)(8.20125,1.1)
\psframe[linewidth=0.04,dimen=outer](4.399375,1.1)(3.399375,0.1)
\usefont{T1}{ptm}{m}{n}
\rput(3.9235938,0.61){\texttt{and}}
\psline[linewidth=0.04cm](2.399375,0.9)(3.399375,0.9)
\psline[linewidth=0.04cm](2.399375,0.3)(3.399375,0.3)
\psframe[linewidth=0.04,dimen=outer](6.399375,1.1)(5.399375,-1.1)
\usefont{T1}{ptm}{m}{n}
\rput(5.913594,0.01){\texttt{or}}
\psline[linewidth=0.04cm](2.399375,-0.3)(5.399375,-0.3)
\psline[linewidth=0.04cm](2.399375,-0.9)(5.399375,-0.9)
\psline[linewidth=0.04cm](4.399375,0.6)(5.399375,0.6)
\psline[linewidth=0.04cm](6.399375,0.0)(7.399375,0.0)
\usefont{T1}{ptm}{m}{n}
\rput(2.0807812,0.91){$a^-$}
\usefont{T1}{ptm}{m}{n}
\rput(2.1307812,0.31){$b^+$}
\usefont{T1}{ptm}{m}{n}
\rput(2.0907812,-0.29){$b^-$}
\usefont{T1}{ptm}{m}{n}
\rput(1.5,-0.89){\texttt{input}$(x^+)$}
\usefont{T1}{ptm}{m}{n}
\rput(7.7207813,0.01){$x^+$}
\end{pspicture} 
}
\caption{The Boolean circuit computing $x^+$ from the clauses $(a \vee \overline b \vee x) \wedge (b \vee x)$. \label{fig3}}
\end{center}
\end{figure}
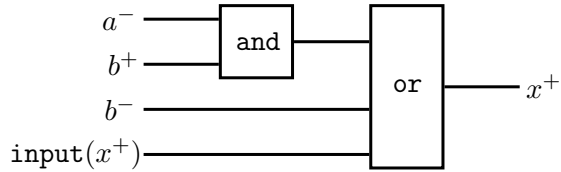

Where $\mathtt{input}(\delta(w)) = x_i^+$ if $w=x_i, 1\leq i \leq n$, $\mathtt{input}(\delta(w)) = x_i^-$ if $w = \overline x_i, 1\leq i \leq n$, and $\mathtt{input}(\delta(w))$ is ignored when $w$ is not related to an input variable (see Figure \ref{fig3} for an example).

Bringing together the circuit parts $C_w$, for any literal $w$ occurring in $\phi$, produces a monotone circuit with loops. These loops (if applicable) are not involved during the unit propagation process because the only deductions they allow can only fix a literal to a value it has already. They can therefore be suppressed by removing the links between any output $u$ of a \texttt{or} gate and any input of a \texttt{and} gate involved in the computation of $u$.

The resulting circuit, which can include unnecessary parts, can simulate any deduction performed by unit propagation on $\phi$ with respect to the values of the input variables $x_1,\ldots,x_n$.

\label{pageend}


\end{spacing}

\bibliographystyle{plain}
\bibliography{grid}

\end{document}